\documentclass{article}

\usepackage[preprint]{neurips_2022}

\usepackage[utf8]{inputenc} 
\usepackage[T1]{fontenc}    
\usepackage{hyperref}       
\usepackage{url}            
\usepackage{booktabs}       
\usepackage{amsfonts}       
\usepackage{nicefrac}       
\usepackage{microtype}      
\usepackage{xcolor}         
\usepackage{subfigure}
\usepackage{graphicx}
\usepackage{threeparttable}
\usepackage{multirow}
\usepackage{CJKutf8}

\title{Oracle-MNIST: a Realistic Image Dataset for Benchmarking Machine Learning Algorithms}

\author{%
  Mei Wang, Weihong Deng \\
  School of Artificial Intelligence\\
  Beijing University of Posts and Telecommunications\\
  \texttt{\{wangmei1, whdeng\}@bupt.edu.cn} \\
}

\begin{document}

\maketitle

\begin{abstract}
  We introduce the Oracle-MNIST dataset, comprising of 28$\times $28 grayscale images of 30,222 ancient characters from 10 categories, for benchmarking pattern classification, with particular challenges on image noise and distortion. The training set totally consists of 27,222 images, and the test set contains 300 images per class. Oracle-MNIST shares the same data format with the original MNIST dataset, allowing for direct compatibility with all existing classifiers and systems, but it constitutes a more challenging classification task than MNIST. The images of ancient characters suffer from 1) extremely serious and unique noises caused by three-thousand years of burial and aging and 2) dramatically variant writing styles by ancient Chinese, which all make them realistic for machine learning research. The dataset is freely available at \url{https://github.com/wm-bupt/oracle-mnist}.
\end{abstract}

\section{Introduction}

In the last few years, fast progress has been unfolding in machine learning (ML) thanks to the release of specialized datasets serving as an experimental testbed and public benchmark of current progress, thus focusing the efforts of the research community. The most widely known dataset in computer vision is the MNIST dataset, which was first introduced in 1998 by \citet{lecun1998gradient}. MNIST is a 10-classs digit classification dataset, and consists of 60,000 grayscale images for training and 10,000 grayscale images for testing. The entire dataset is relatively small, free to access and use, and is encoded and stored in an entirely straightforward manner, which have almost certainly contributed to its widespread use.

However, with the discovery of improved learning algorithms, the performance has been saturated on MNIST. For example, Convolutional Neural Networks (CNNs) \citep{krizhevsky2012imagenet,he2016deep} can easily achieve an accuracy of above 99\%. \emph{This is partially attributed to the benchmark that does not capture requirements of many real-world scenarios.} 
To avoid the saturated performance and offer challenges for the improved ML algorithms, some modified MNIST datasets are constructed, e.g., EMNIST \citep{cohen2017emnist} and Fashion-MNIST \citep{xiao2017fashion}. EMNIST extends the number of classes by introducing uppercase and lowercase characters, but the extra classes require a change of the framework of deep neural network used by MNIST. Fashion-MNIST contains 70,000 grayscale images of 10-class fashion products. These product images are taken from Zalando’s website\footnote{Zalando is the Europe’s largest online fashion platform. http://www.zalando.com} shot by professional photographers, and thus are clear and standardized. However, it fails to capture as wide of a range of variations as possible in the real world.

The purpose of this paper is to provide a realistic and challenging dataset, called Oracle-MNIST, to facilitate easy and fast evaluation for ML algorithms on the real-world images of ancient characters. Oracle-MNIST contains 30,222 images of oracle characters belonging to 10 categories.

\begin{enumerate}
  \item \textbf{Real-world challenge.} Different from handwritten digits, oracle characters are scanned from the real oracle-bone surface. Therefore, Oracle-MNIST suffers from extremely serious and unique noises caused by thousands of years of burial and aging, and contains various writing styles in each category which all make it more realistic and difficult for ML research.
  \item \textbf{Ease-of-use.} Following the original MNIST, the images in Oracle-MNIST have 28$\times $28 grayscale pixels. It can immediately compatible with any ML package capable of working with the MNIST dataset since it shares the same data format. In fact, the only change one needs to use this dataset is to change the URL from where the MNIST dataset is fetched.
\end{enumerate}

\begin{figure}
\centering
\subfigure[Example of oracle bone]{
\label{sensi_b} 
\includegraphics[height=3.5cm]{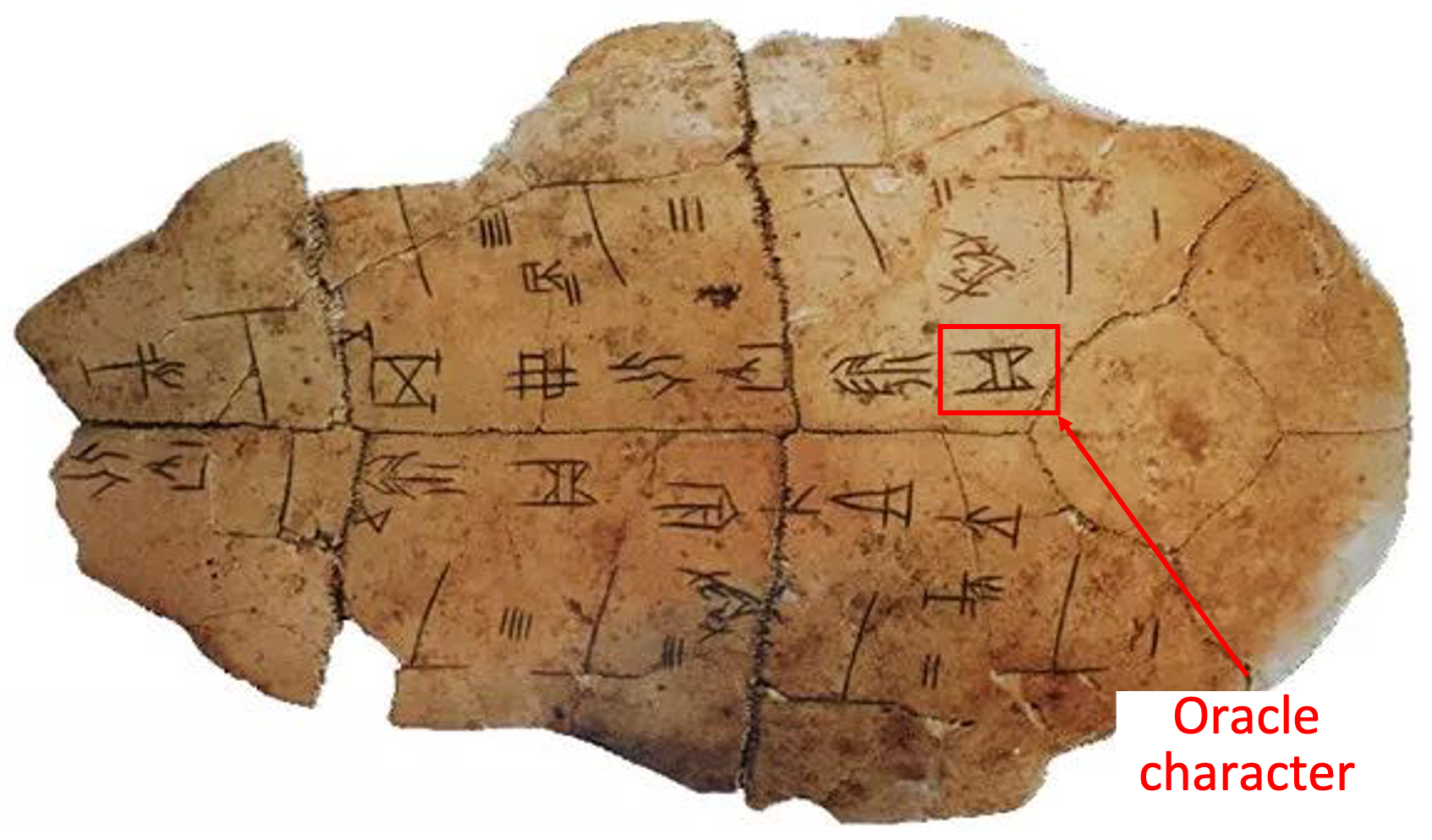}}
\hspace{0.5cm}
\subfigure[Evolution of Chinese character]{
\label{sensi_b} 
\includegraphics[height=3.5cm]{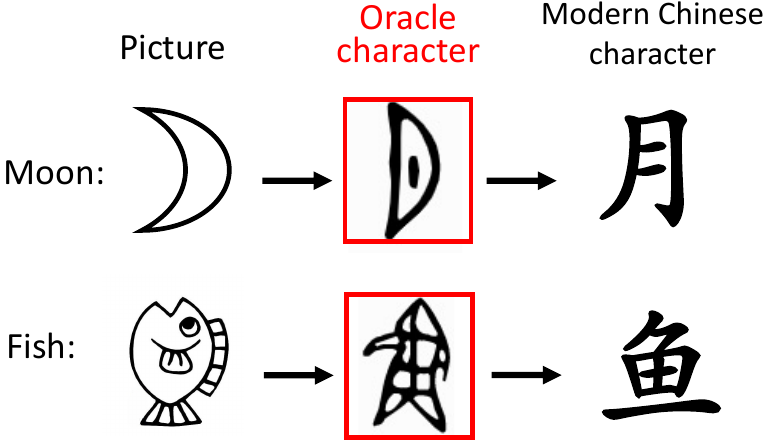}}
\caption{Oracle characters are the oldest hieroglyphs in China, which were inscribed on (a) oracle bones about 3000 years ago. (b) Despite the pictorial nature of oracle characters, it constitutes a fully functional and well-developed writing system.}
\label{oracle-bone} 
\end{figure}

\section{Oracle-MNIST Dataset}

\subsection{Discovery of Oracle Characters} \label{oracle}

Ancient history relies on the study of ancient characters. As the oldest hieroglyphs in China, oracle characters \citep{flad2008divination,keightley1997graphs}, with a history spanning nearly three millennia, have contributed greatly to modern civilization, enabling the Chinese culture to be passed on from generation to generation and become the only civilization to last up to the present. As shown in Fig. \ref{oracle-bone}, oracle characters are engraved on tortoise shells and animal bones, and record the life and history of the Shang Dynasty (around 1600-1046 B.C.), including divination practices, war expeditions, hunting, medical treatments, and childbirth. They were first discovered by a merchant called Wang Xirong in 1899, during the Qing Dynasty (1644-1911). In the early 20th century, Chinese researchers excavated numerous oracle bones at Xiaotun Village in Anyang, Henan Province, capital of the Shang Dynasty. Since then, the research on oracle characters has attracted much attention. It is of vital importance for Chinese etymologies and calligraphy as well as learning the culture and history of ancient China and even the world.

\begin{figure}[b]
\centering
\subfigure[Scanned inscription]{
\label{scan1} 
\includegraphics[height=3.5cm]{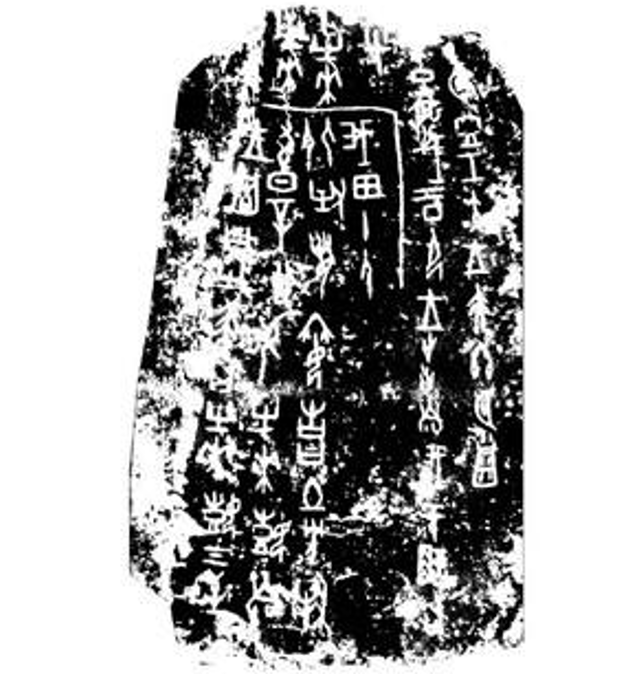}}
\hspace{0.1cm}
\subfigure[Character: ‘horse’]{
\label{scan2} 
\includegraphics[height=3.3cm]{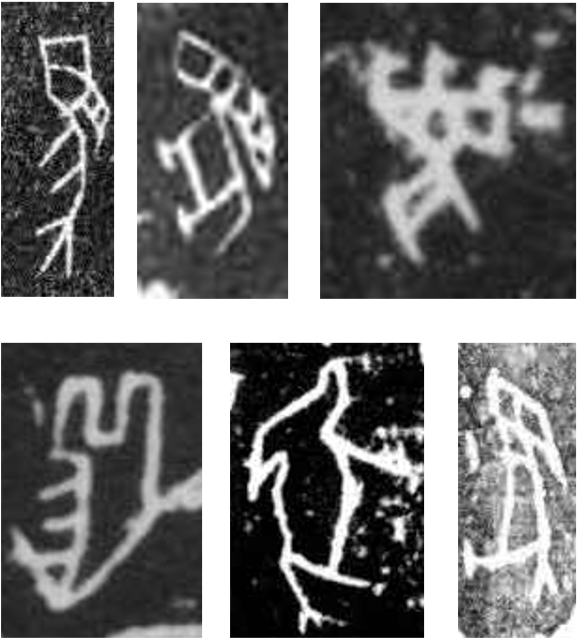}}
\hspace{0.4cm}
\subfigure[Character: ‘wood’]{
\label{scan3} 
\includegraphics[height=3.3cm]{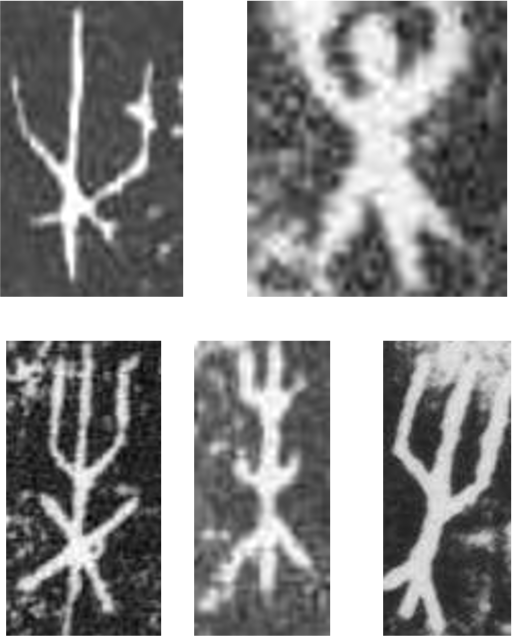}}
\hspace{0.4cm}
\subfigure[Character: ‘cattle’]{
\label{scan4} 
\includegraphics[height=3.3cm]{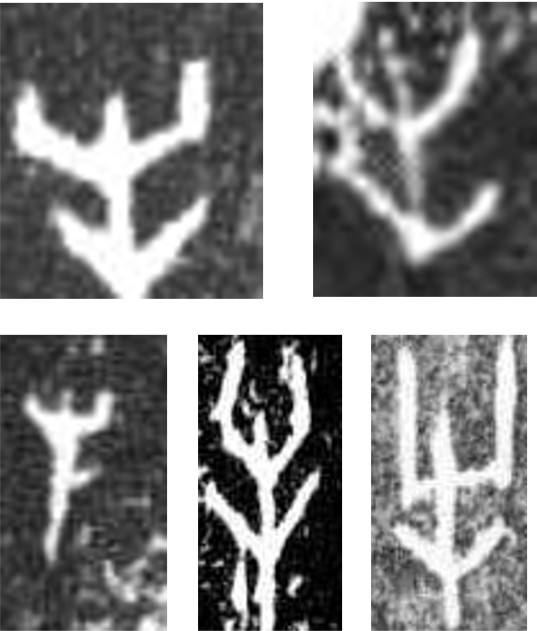}}
\caption{(a) Example of scanned oracle inscription. (b-d) Examples of scanned oracle characters. Different writing styles lead to a high degree of intra-class variance and inter-class similarity. }
\label{scan} 
\end{figure}

Most of oracle characters are stored by scanned images, which are generated by reproducing the oracle-bone surface by placing a piece of paper over the subject and then rubbing the paper with rolled ink, as shown in Fig. \ref{scan1}. Recognizing these oracle characters is difficult for both experts and machines. Thus far, nearly 4,500 different oracle characters have been discovered, but only about 2,200 characters have been successfully deciphered. The reasons are as follows. (1) \textbf{Abrasion and noise.} Many oracle-bone inscriptions have been damaged over the centuries and their texts are now fragmentary. The aging process has also made the inscriptions less legible so that these scanned characters are broken and contain serious noises. (2) \textbf{Large variance.} Different writing styles lead to a high degree of intra-class variance. Characters belonging to the same category largely vary in stroke and even topology, as shown in Fig. \ref{scan2}. Some characters belonging to the different categories are similar to each other, which brings great difficulty for recognition. For example, the characters of `wood' and `cattle' categories only differ in some small details shown in Fig. \ref{scan3} and \ref{scan4}.

\subsection{Details of Dataset} \label{dataset}

\begin{figure}
\centering
\includegraphics[height=4cm]{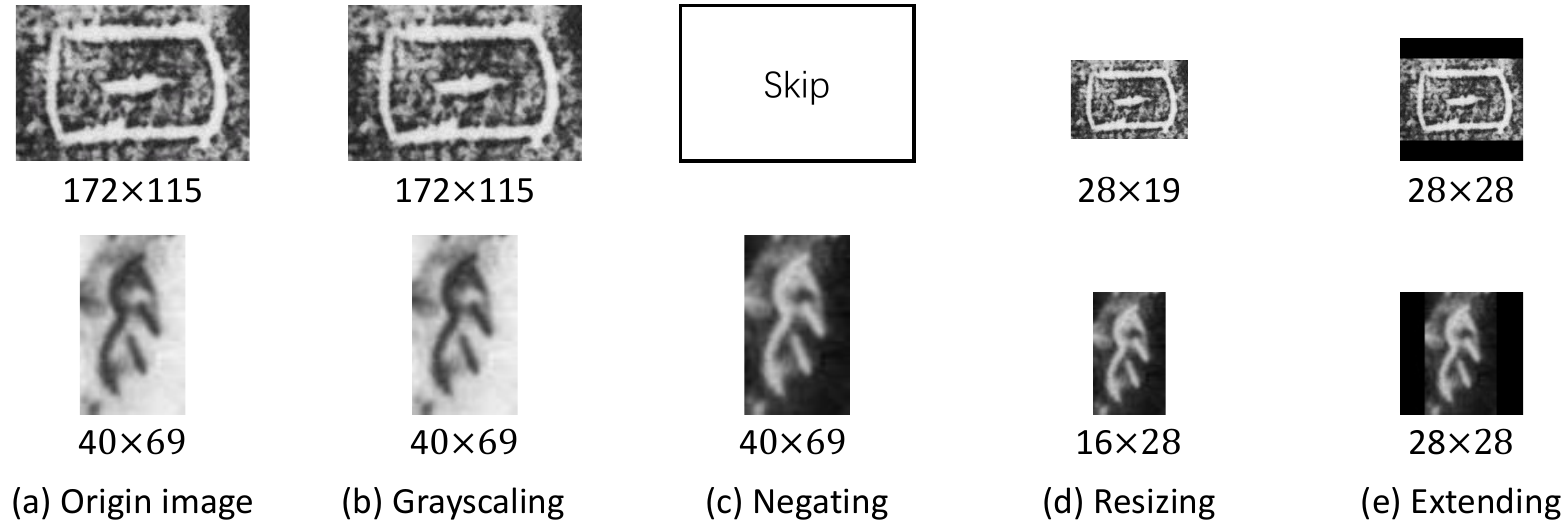}
\caption{Diagram of the conversion process used to generate Oracle-MNIST dataset. Two examples from `sun' and `not' categories are depicted, respectively. }
\label{convert} 
\end{figure}

\begin{table}
  \caption{Files contained in the Oracle-MNIST dataset.}
  \label{file}
  \centering
  \begin{threeparttable}
  \begin{tabular}{llll}
    \toprule
    Name     & Description     &  \#Images & Size \\
    \midrule
    train-images-idx3-ubyte.gz & Training set images  & 27,222   & 12.4 MBytes  \\
    train-labels-idx1-ubyte.gz     & Training set labels & 27,222   & 13.7 KBytes   \\
    t10k-images-idx3-ubyte.gz\tnote{1}     & Test set images       & 3,000 & 1.4 MBytes \\
    t10k-labels-idx1-ubyte.gz\tnote{1} & Test set labels & 3,000 & 1.6 KBytes\\
    \bottomrule
    \end{tabular}
    \begin{tablenotes}
     \item[1] Although our test set consists of only 3K images, it is called `t\textbf{10}k' instead of `t\textbf{3}k' to be consistent with the original MNIST dataset such that it can be easily compatible with any ML package.
    \end{tablenotes} 
    \end{threeparttable}
\end{table} 

Oracle-MNIST is based on the collection of YinQiWenYuan website\footnote{YinQiWenYuan is a large oracle-bone platform (http://jgw.aynu.edu.cn/ajaxpage/home2.0) constructed by AnYang Normal University.}. Those oracle characters are scanned from the real oracle-bone surface, thus they are broken and suffer from serious noises. Each scanned image is centered by one single character. Most of the original images have gray or black backgrounds and vary in resolution.

We selected 30,222 commonly-used characters of 10 classes to build Oracle-MNIST. The original images are then fed into the following conversion pipeline, which is visualized in Fig. \ref{convert}. We also attempt to process the images by some image enhancement techniques, e.g., gray stretch and histogram equalization. Although the visual quality of images is successfully improved, the recognition performance would slightly degrade. Therefore, no image enhancement technology is applied to Oracle-MNIST. We also make the original images available and left the data processing job to the algorithm developers.

\begin{enumerate}
  \item Converting the image to 8-bit grayscale pixels.
  \item Negating the intensities of the image if its foreground is darker than the background.
  \item Resizing the longest edge of the image to 28 using a bi-cubic interpolation algorithm.
  \item Extending the shortest edge to 28 and put the image to the center of the canvas.
\end{enumerate}

We utilize the meanings of characters as their class labels. The labels are manually annotated by experts in archeology or paleography. Table \ref{some example} gives a summary of all class labels in Oracle-MNIST with examples for each class.

Finally, we divide the dataset into a training and a test set, and make sure that they are disjoint. The training set totally consists of randomly-selected 27,222 images, and the test set contains 300 images per class. Images and labels are stored in the same file format as the MNIST dataset, which is designed for storing vectors and multidimensional matrices. The result files are listed in Table \ref{file}.

\begin{table}[h]
\caption{Class labels, example images and the number of training images in Oracle-MNIST dataset.}
\label{some example}
  \centering
  \setlength{\tabcolsep}{1.3mm}{
  \begin{threeparttable}
  \begin{tabular}{llll}
    \toprule
    \multirow{2}{*}{Label} & Description  &\multirow{2}{*}{\makebox[0.72\textwidth][l]{Examples}} \\ 
    & (\scriptsize{\# Training images})  & \\ \midrule
    0 & big \begin{CJK}{UTF8}{gkai}大\end{CJK}(2433) & \begin{minipage}{0\columnwidth}
		\centering
		\raisebox{-.8\height}{\includegraphics[height=1.35cm]{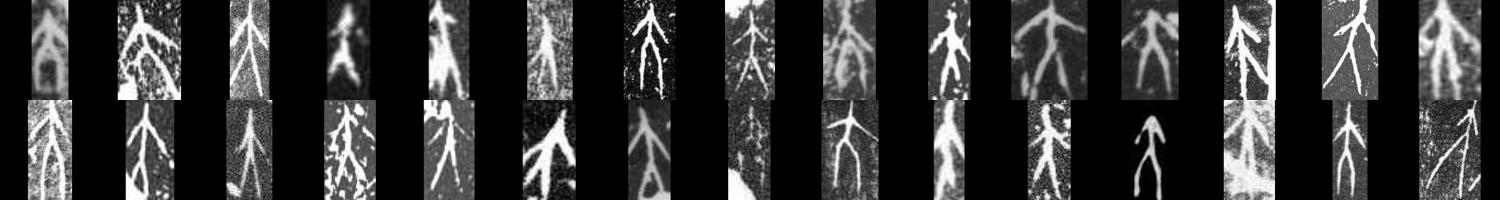}}
	\end{minipage}\\ \midrule
    1 & sun \begin{CJK}{UTF8}{gkai}日\end{CJK}(2765) & \begin{minipage}{0\columnwidth}
		\centering
		\raisebox{-.8\height}{\includegraphics[height=1.35cm]{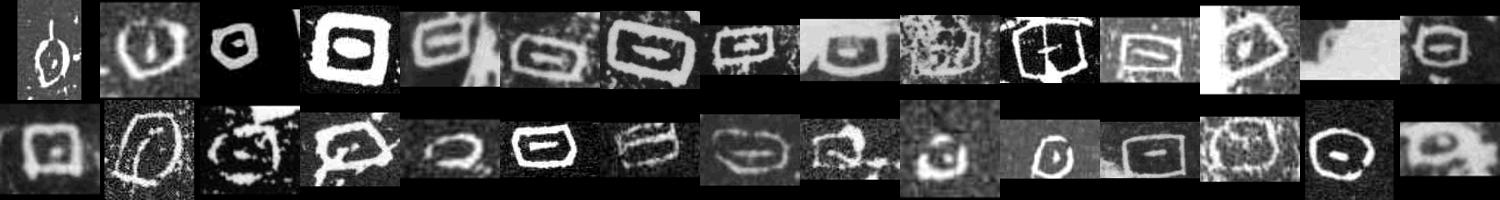}}
	\end{minipage}\\ \midrule
    2 & moon \begin{CJK}{UTF8}{gkai}月\end{CJK}(2668) & \begin{minipage}{0\columnwidth}
		\centering
		\raisebox{-.8\height}{\includegraphics[height=1.35cm]{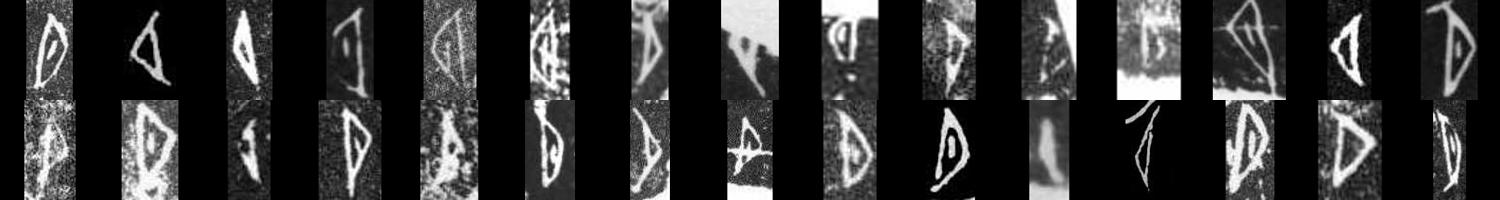}}
	\end{minipage}\\ \midrule
    3 & cattle \begin{CJK}{UTF8}{gkai}牛\end{CJK}(2614) & \begin{minipage}{0\columnwidth}
		\centering
		\raisebox{-.8\height}{\includegraphics[height=1.35cm]{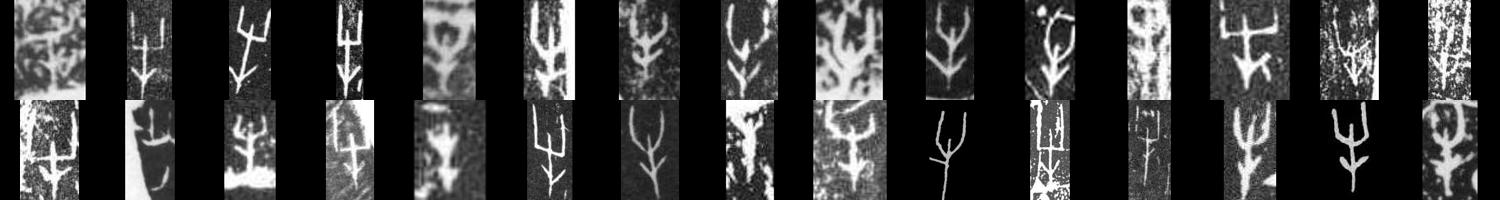}}
	\end{minipage}\\ \midrule
    4 & next \begin{CJK}{UTF8}{gkai}翌\end{CJK}(2610) & \begin{minipage}{0\columnwidth}
		\centering
		\raisebox{-.8\height}{\includegraphics[height=1.35cm]{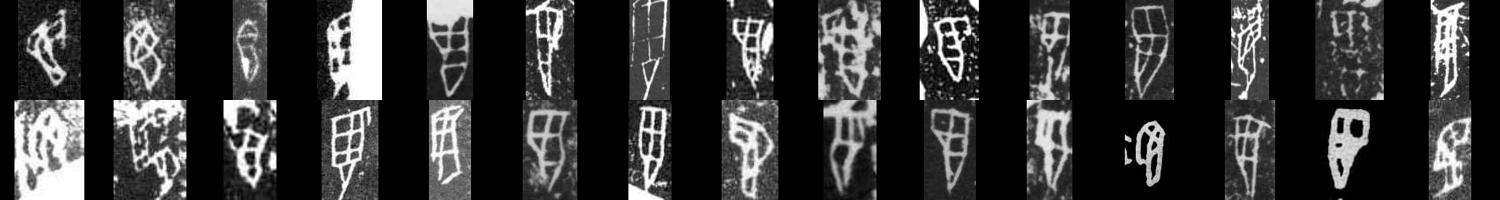}}
	\end{minipage}\\ \midrule
    5 & field \begin{CJK}{UTF8}{gkai}田\end{CJK}(2328) & \begin{minipage}{0\columnwidth}
		\centering
		\raisebox{-.8\height}{\includegraphics[height=1.35cm]{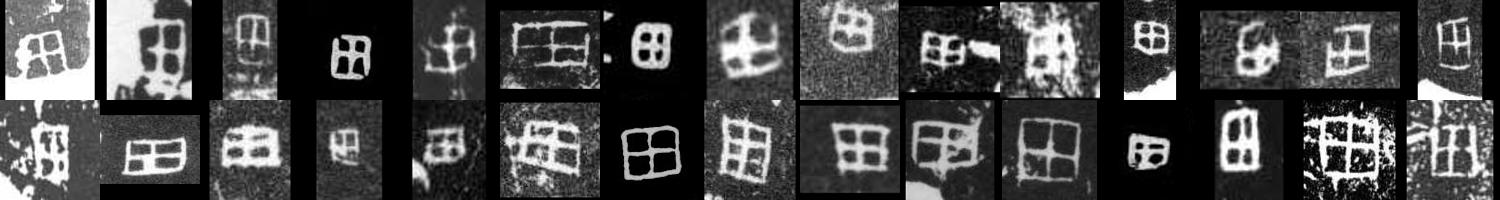}}
	\end{minipage}\\ \midrule
    6 & not \begin{CJK}{UTF8}{gkai}勿\end{CJK}(2710) & \begin{minipage}{0\columnwidth}
		\centering
		\raisebox{-.8\height}{\includegraphics[height=1.35cm]{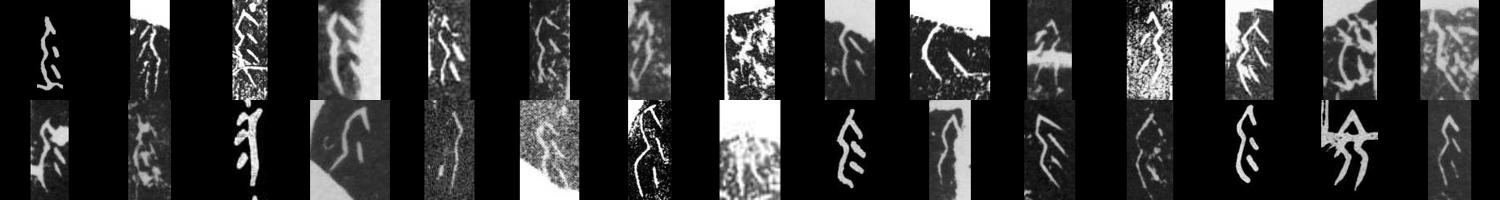}}
	\end{minipage}\\ \midrule
    7 & arrow \begin{CJK}{UTF8}{gkai}矢\end{CJK}(2360) & \begin{minipage}{0\columnwidth}
		\centering
		\raisebox{-.8\height}{\includegraphics[height=1.35cm]{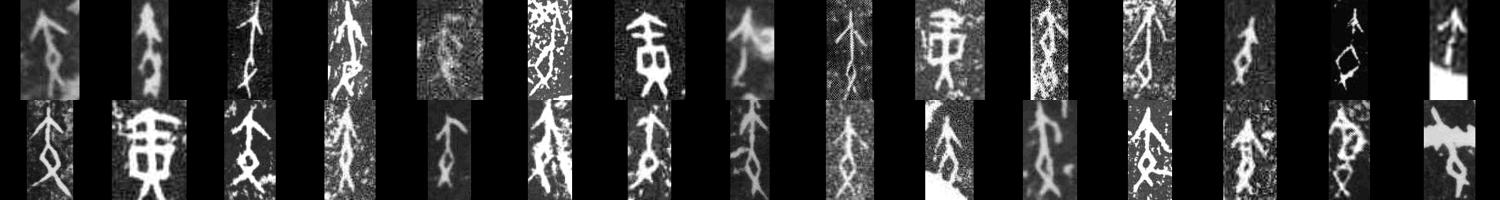}}
	\end{minipage}\\ \midrule
    8 & time\tnote{1} \begin{CJK}{UTF8}{gkai}巳\end{CJK}(3335) & \begin{minipage}{0\columnwidth}
		\centering
		\raisebox{-.8\height}{\includegraphics[height=1.35cm]{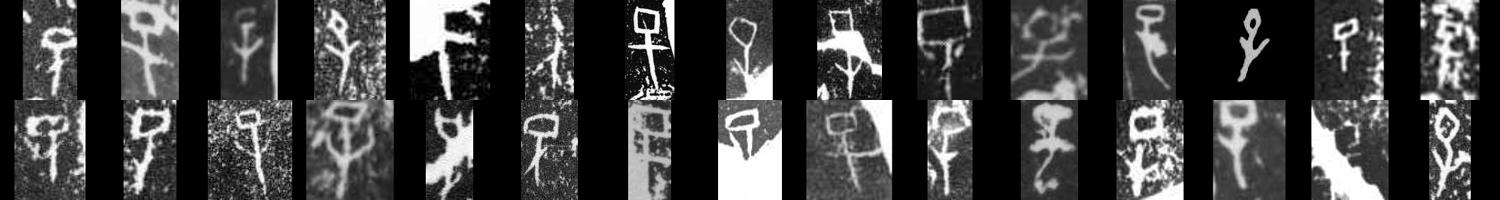}}
	\end{minipage}\\ 
    9 & wood \begin{CJK}{UTF8}{gkai}木\end{CJK}(3399) & \begin{minipage}{0\columnwidth}
		\centering
		\raisebox{-.8\height}{\includegraphics[height=1.35cm]{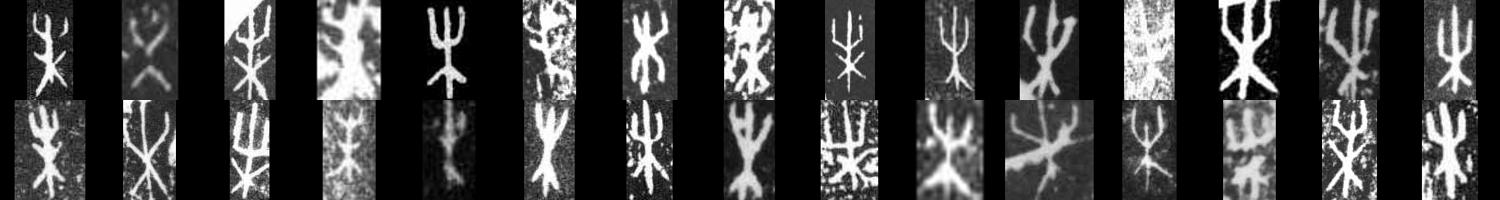}}
	\end{minipage}\\ 
    \bottomrule
  \end{tabular}
  \begin{tablenotes}
     \item[1] 9-11 a.m. (one of the Earthly Branches which are ancient China’s systems for keeping time).
    \end{tablenotes} 
    \end{threeparttable}}
\end{table}

\section{Experiments}

We evaluate some algorithms with different parameters on Oracle-MNIST and report the results in Table \ref{result}. For each algorithm, the average classification accuracy is reported based on three repeated experiments. The benchmarks on the MNIST and Fashion-MNIST dataset are also included for a side-by-side comparison. 

From the results, we have the following observations. First, classic (shallow) ML algorithms can easily achieve 97\% on the MNIST dataset which proves that MNIST is too easy to evaluate the algorithms. Our Oracle-MNIST dataset provides 10-class images of ancient characters and further captures as wide of a range of variations as possible in the real world to pose a more challenging classification task than the MNIST digits data and Fashion-MNIST data. As we can see that all classic (shallow) ML algorithms perform the best on MNIST, followed by Fashion-MNIST, and the worst on Oracle-MNIST. For example, the random forest classifier achieves the accuracies of 97.1\% , 87.1\% and 64.9\%, respectively. This is because a high degree of intra-class variance and inter-class similarity, described above in Section  \ref{oracle}, would bring great difficulty for classification. Moreover, the scanned oracle images are seriously degraded and even completely lost their discriminative glyph information caused by blur, noise and occlusion. 

Second, CNN outperforms all of the classic (shallow) ML algorithms on Oracle-MNIST. Benefitting from local receptive fields and spatial or temporal subsampling, CNN can force the extraction of local features and reduce the sensitivity of the output to shifts and distortions \citep{lecun1995convolutional}. Therefore, the real-world challenges, e.g., different writing styles, noise and occlusion, can be tackled to some extent. However, the performance on Oracle-MNIST has not been saturated. The CNN utilized in this paper achieves an error rate of 6.2\% on Oracle-MNIST, and there is still room for improvement. Inspite of the powerful representation ability of CNN, the problem of recognizing these ancient characters remains to be fully solved.

\begin{table}[h]
  \caption{Benchmark on Oracle-MNIST, Fashion-MNIST and MNIST.}
  \label{result}
  \centering
   \setlength{\tabcolsep}{1.2mm}{
  \begin{tabular}{llccc}
    \toprule
    \multirow{2}{*}{Algorithm}     &\makebox[0.42\textwidth][l]{ \multirow{2}{*}{Parameter}}    &  \multicolumn{3}{c}{Test Accuracy} \\ \cmidrule(r){3-5}
    & & Oracle & Fashion & MNIST \\
    \midrule
    CNN & \scriptsize{2$\times $Conv-Pool-ReLu, 2$\times $FC,  Dropout} & 93.8 & 92.1 & 99.3	\\
    & \scriptsize{2$\times $Conv-Pool-ReLu, 2$\times $FC} & 92.8 & 90.8 & 99.4 \\
    & 	\scriptsize{1$\times $Conv-Pool-ReLu, 2$\times $FC} & 91.6 & 91.2 & 99.2 \\	
    \midrule
    SVC & \scriptsize{C=10, kernel=rbf} & 75.5 & 89.7 & 97.3	\\
    & \scriptsize{C=100, kernel=rbf} & 75.0 & 89.0 & 97.2 \\
    & \scriptsize{C=100, kernel=poly} & 74.5 & 89.0 & 97.8  \\	
    & \scriptsize{C=10, kernel=poly} & 73.2 & 89.1 & 97.6 \\
    & \scriptsize{C=1, kernel=rbf} &  71.3 & 87.9 & 96.6 \\
    & \scriptsize{C=1, kernel=poly} & 62.9 & 87.3 & 95.7 \\
    & \scriptsize{C=1, kernel=linear} & 57.6 & 83.9 & 92.9 \\
    & \scriptsize{C=10, kernel=linear} & 56.7 & 82.9 &  92.7 \\
    & \scriptsize{C=100, kernel=linear} & 56.2 & 82.7 &  92.6 \\
     \midrule
    MLPClassifier & \scriptsize{activation=relu, hidden\_layer\_sizes=[100]} & 74.7 & 87.1 & 	97.2	\\
    & \scriptsize{activation=relu, hidden\_layer\_sizes=[100, 10]} & 72.6 & 87.0 & 97.2	 \\
    & \scriptsize{activation=tanh, hidden\_layer\_sizes=[100]} & 66.7 & 86.8 & 	96.2	 \\
    & \scriptsize{activation=tanh, hidden\_layer\_sizes=[100, 10]} & 65.5 & 86.3 & 95.7 \\	
    & \scriptsize{activation=relu, hidden\_layer\_sizes=[10, 10]} & 61.2 & 85.0 & 	93.6	\\
    & \scriptsize{activation=relu, hidden\_layer\_sizes=[10]}	 & 60.7 & 84.8 & 93.3 \\
    & \scriptsize{activation=tanh, hidden\_layer\_sizes=[10]} & 58.4 & 84.1 & 92.1 \\	
    & \scriptsize{activation=tanh, hidden\_layer\_sizes=[10, 10]} & 58.4 & 84.0 & 92.1 \\	
    \midrule
    GradientBoostingClassifier & \scriptsize{n\_estimators=100, loss=deviance, max\_depth=10} &  72.5 & 88.0 & 96.9	\\
    & \scriptsize{n\_estimators=50, loss=deviance, max\_depth=10} & 69.9 & 87.2 & 96.4 \\
    & \scriptsize{n\_estimators=100, loss=deviance, max\_depth=3} & 69.7 & 86.2 & 94.9 \\		
    \midrule
    & & \multicolumn{3}{c}{Continued on next page} \\
    \bottomrule
    \end{tabular}}
\end{table} 

\begin{table}
\renewcommand{\thetable}{3}
  \caption{continued from previous page.}
  \label{result}
  \centering
   \setlength{\tabcolsep}{1.2mm}{
  \begin{tabular}{llccc}
    \toprule
    \multirow{2}{*}{Algorithm}     & \multirow{2}{*}{Parameter}    &  \multicolumn{3}{c}{Test Accuracy} \\ \cmidrule(r){3-5}
    & & Oracle & Fashion & MNIST \\
        \midrule
    & \scriptsize{n\_estimators=50, loss=deviance, max\_depth=3} & 64.6 & 84.0 & 92.6	\\
    & \scriptsize{n\_estimators=10, loss=deviance, max\_depth=10} &  59.9 & 84.9 &  93.3 \\
        \midrule
    RandomForestClassifier     & \scriptsize{n\_estimators=100, criterion=gini, max\_depth=100} & 65.0 & 87.2 &  97.0 \\	
    & \scriptsize{n\_estimators=100, criterion=entropy, max\_depth=50} & 65.0 & 87.2 & 96.9 \\	
    & \scriptsize{n\_estimators=100, criterion=gini, max\_depth=50} & 64.9 & 87.1 & 97.1 \\	
    & \scriptsize{n\_estimators=100, criterion=entropy, max\_depth=100} & 64.9 & 87.3 & 	97.0 \\	
    & \scriptsize{n\_estimators=50, criterion=gini, max\_depth=100} & 63.6 & 86.9 & 96.7 \\	
    & \scriptsize{n\_estimators=50, criterion=entropy, max\_depth=50}	 & 62.9 & 87.1 & 96.7 \\	
    & \scriptsize{n\_estimators=50, criterion=gini, max\_depth=50} & 62.6 & 87.0 & 96.8 \\	
    & \scriptsize{n\_estimators=50, criterion=entropy, max\_depth=100} & 62.5 & 87.2 & 96.8 \\	
    & \scriptsize{n\_estimators=100, criterion=gini, max\_depth=10} & 58.3 & 83.5 & 	94.9	\\
    & \scriptsize{n\_estimators=100, criterion=entropy, max\_depth=10} & 58.3 & 83.8 & 95.0 \\
    & \scriptsize{n\_estimators=50, criterion=entropy, max\_depth=10}	 & 58.0	 & 83.8 & 94.7	\\
    & \scriptsize{n\_estimators=50, criterion=gini, max\_depth=10} & 57.6 & 83.4 & 	94.5	\\
    & \scriptsize{n\_estimators=10, criterion=gini, max\_depth=10} & 53.2 & 82.5 & 	93.0	\\
    & \scriptsize{n\_estimators=10, criterion=entropy, max\_depth=10}	 & 52.8 & 82.8 & 	93.3	\\
    & \scriptsize{n\_estimators=10, criterion=entropy, max\_depth=100} & 52.1 & 85.2 & 	94.9	\\
    & \scriptsize{n\_estimators=10, criterion=gini, max\_depth=100} & 52.0 & 84.7 & 	94.8	\\
    & \scriptsize{n\_estimators=10, criterion=entropy, max\_depth=50}	 & 51.8 & 85.3 & 	94.9	\\
    & \scriptsize{n\_estimators=10, criterion=gini, max\_depth=50} & 51.3 & 84.8 & 	94.8	\\
    \midrule
    KNeighborsClassifier & \scriptsize{weights=distance, n\_neighbors=9, p=2} & 62.7 & 84.9 & 	94.4	\\
    & \scriptsize{weights=distance, n\_neighbors=9, p=1} & 61.8 & 85.4 & 	95.5	\\
    & \scriptsize{weights=uniform, n\_neighbors=9, p=1} & 61.6 & 85.3 & 	95.5	\\
    & \scriptsize{weights=uniform, n\_neighbors=9, p=2} & 61.5 & 84.7  & 94.3 \\
    & \scriptsize{weights=distance, n\_neighbors=5, p=1} & 60.3 & 85.4 & 	95.9	\\
    & \scriptsize{weights=uniform, n\_neighbors=5, p=1} & 59.6 & 85.2 & 	95.7	\\
    & \scriptsize{weights=distance, n\_neighbors=5, p=2} & 59.5 & 85.2 & 	94.5	\\
    & \scriptsize{weights=uniform, n\_neighbors=5, p=2} & 59.0 & 84.9 & 	94.4	\\
    & \scriptsize{weights=uniform, n\_neighbors=1, p=1} & 55.8 & 83.8 & 	95.5	\\
    & \scriptsize{weights=distance, n\_neighbors=1, p=1} & 55.8 & 83.8 & 	95.5	\\
    & \scriptsize{weights=distance, n\_neighbors=1, p=2} & 55.7 & 83.9 & 	94.3	\\
    & \scriptsize{weights=uniform, n\_neighbors=1, p=2} & 55.7 & 83.9 & 	94.3	\\
     \midrule
    LogisticRegression     & \scriptsize{C=10, multi\_class=ovr, penalty=l2} & 59.8 & 83.9 & 91.6 \\
    & \scriptsize{C=100, multi\_class=ovr, penalty=l2} & 59.8 & 83.6 & 91.6 \\
    & \scriptsize{C=1, multi\_class=ovr, penalty=l2} & 59.7 & 84.1 & 91.7 \\
    \midrule
    LinearSVC & \scriptsize{loss=hinge, C=1, multi\_class=crammer\_singer, penalty=l2} & 58.1 & 83.5 & 	91.9	\\
    & \scriptsize{loss=squared\_hinge, C=1, multi\_class=crammer\_singer, penalty=l2} & 58.0 & 83.4 & 91.9	\\
    & \scriptsize{loss=hinge, C=1, multi\_class=crammer\_singer, penalty=l1} & 57.8 & 83.3 & 	91.9	 \\
    & \scriptsize{loss=squared\_hinge, C=1, multi\_class=crammer\_singer, penalty=l1} & 57.3 & 83.3 & 	91.9	\\
    & \scriptsize{loss=squared\_hinge, C=1, multi\_class=ovr, penalty=l2} & 55.8	 & 82.0 & 91.2	\\
    & \scriptsize{loss=hinge, C=1, multi\_class=ovr, penalty=l2} & 54.8 & 83.6 & 	91.7	\\
     \midrule
    SGDClassifier & \scriptsize{loss=log, penalty=l1} & 56.7 & 81.5 & 	91.0 \\
    & \scriptsize{loss=log, penalty=elasticnet } & 56.0 & 81.6 & 	91.2	\\
    & \scriptsize{loss=hinge, penalty=l1} & 55.6 & 81.5 & 91.1 \\
    & \scriptsize{loss=log, penalty=l2}	 & 55.1 & 81.3 & 	91.3	\\
    & \scriptsize{loss=hinge, penalty=l2} & 54.9 & 81.9 & 	91.4	\\
    & \scriptsize{loss=hinge, penalty=elasticnet} & 54.7 & 81.6 & 	91.3	\\
    & \scriptsize{loss=modified\_huber, penalty=elasticnet} & 51.5 & 81.3 & 	91.4	\\
    & \scriptsize{loss=modified\_huber, penalty=l1} & 50.7 & 81.7 & 	91.0 \\
    & \scriptsize{loss=perceptron, penalty=l1} & 49.4 & 81.8 & 	91.2	\\
    & \scriptsize{loss=modified\_huber, penalty=l2}  & 49.0 & 81.6 &  	91.3	\\

    \midrule
    & & \multicolumn{3}{c}{Continued on next page} \\
     \bottomrule
    \end{tabular}}
\end{table}

\begin{table}
\renewcommand{\thetable}{3}
  \caption{continued from previous page.}
  \label{result}
  \centering
   \setlength{\tabcolsep}{1.2mm}{
  \begin{tabular}{llccc}
    \toprule
    \multirow{2}{*}{Algorithm}     & \multirow{2}{*}{Parameter}    &  \multicolumn{3}{c}{Test Accuracy} \\ \cmidrule(r){3-5}
    & & Oracle & Fashion & MNIST \\
        \midrule
    \makebox[0.25\textwidth][l]{}& \makebox[0.45\textwidth][l]{\scriptsize{loss=perceptron, penalty=l2}} & 47.9 & 81.4 & 	91.3	\\
    & \scriptsize{loss=perceptron, penalty=elasticnet} & 47.4 & 81.4 & 	91.2	\\
    & \scriptsize{loss=squared\_hinge, penalty=l1  } & 46.5	& 81.3 & 91.1	\\
    & \scriptsize{loss=squared\_hinge, penalty=l2	}& 45.0 & 81.4 & 	91.2	\\
    & \scriptsize{loss=squared\_hinge, penalty=elasticnet}	 & 42.1 & 81.5 & 	91.4	\\
     \bottomrule
    \end{tabular}}
\end{table} 

\section{Conclusions}

In this paper, we present a realistic and challenging benchmark dataset, i.e. Oracle-MNIST. It contains 28$\times $28 grayscale images of 30,222 ancient characters belonging to 10 categories, for benchmarking the robustness to image noise and distortion in computer vision. Oracle-MNIST is converted to a format that is directly compatible with classifiers built to handle the MNIST dataset. With thousands of years of burial and aging, these ancient characters suffer from noise, and additionally vary in writing styles, which bring great difficulty for recognition. Benchmark results show that the classification task of ancient characters is indeed more challenging.

\small
\bibliography{egbib}
\bibliographystyle{iclr2022}

\end{document}